\documentclass[a4paper]{article}
\let\svthefootnote\thefootnote
\usepackage[english]{babel}
\usepackage[utf8x]{inputenc}
\usepackage[T1]{fontenc}
\usepackage{url}

\usepackage{color}
\usepackage{soul}

\def\citeA#1{\citet{#1}}
\def\citeNP#1{\citealp{#1}}

\usepackage{natbib}
\bibpunct{(}{)}{,}{a}{}{;}

\usepackage{color,soul}
\usepackage{amsmath}
\usepackage{graphicx}
\usepackage[colorinlistoftodos]{todonotes}
\usepackage{ulem}

\title{Analyzing and Interpreting \\Neural Networks for NLP: \\ A Report on the First BlackboxNLP Workshop}

\author{Afra Alishahi\\Tilburg University \and 
Grzegorz Chrupała\\Tilburg University \and  Tal Linzen\\
Johns Hopkins University}
\date{}
\begin{document}
\maketitle
\let\thefootnote\relax
\footnote{Authors contributed equally and are listed in alphabetical order.}\addtocounter{footnote}{-1}\let\thefootnote\svthefootnote

\begin{abstract}
The EMNLP 2018 workshop BlackboxNLP was dedicated to resources and techniques specifically developed for analyzing and understanding the inner-workings and representations acquired by neural models of language. Approaches included: systematic manipulation of input to neural networks and investigating the impact on their performance, testing whether interpretable knowledge can be decoded from intermediate representations acquired by neural networks, proposing modifications to neural network architectures to make their knowledge state or generated output more explainable, and examining the performance of networks on simplified or formal languages. Here we review a number of representative studies in each category.
\end{abstract}

\section{Introduction}

Neural networks have rapidly become a central component in language and speech understanding systems in the last few years. The improvements in accuracy and performance brought by the introduction of neural networks have typically come at the cost of our understanding of the system: what are the representations and computations that the networks learn? 

In October 2018, we organized a workshop called BlackboxNLP\footnote{\url{https://blackboxnlp.github.io/2018/}} as part of the Empirical Methods in Natural Language Processing (EMNLP 2018) conference to bring together researchers who are attempting to peek inside the neural network black box, taking inspiration from machine learning, psychology, linguistics and neuroscience. The workshop generated a lot of interest withing the NLP community: we received a total of 75 submissions, and more than 600 EMNLP attendees signed up for the workshop. The topics presented and discussed in the workshop were diverse, representing the variety of methodologies, resources and techniques currently being used for analyzing the inner-workings and knowledge acquired by neural networks. The following themes, however, were dominant:
\begin{itemize}
    \item Systematic manipulation of  input to neural networks and investigating the impact on their performance, often through developing annotated and specialized datasets;
    \item Testing whether interpretable knowledge can be decoded from intermediate representations acquired by neural networks, often through passing them to {\it diagnostic classifiers} or other downstream tasks;
    \item Proposing modifications to neural network architectures to make their knowledge state or generated output more explainable;
    \item Examining the performance of the network on simplified or formal languages.
\end{itemize}
In this report, will briefly review some representative studies in each category, and provide some thoughts for future research.

\section{Input manipulation}
\label{sec:dataset}

A number of studies focus on the impact of input on the performance of the model, and use this as a diagnostic method for identifying the important characteristics that affect the model's decisions. 

\subsection{Simple input changes}
A straightforward approach applies simple preprocessing techniques to the training data of a model, and examines how the performance of the model changes: \citeA{W18-5404} focus on the impact of punctuation on syntactic parsing, and show that neural dependency parsers are more sensitive to punctuation-free input than the previous generation of parsers. In a similar vein, \citeA{W18-5406} investigate the impact of various text preprocessing techniques, such as tokenizing, lemmatizing and chunking, on the performance of neural models of text classification and sentiment analysis, and suggest preprocessing guidelines for training word embeddings.

\subsection{Datasets}
Alternatively, some develop diagnostic resources containing carefully constructed input examples which vary in their degree of (linguistic) complexity or in providing different types of challenges to a model. Diverse NLI Collection (DNC; \citeNP{W18-5441}) is such a case, which presents a large collection of datasets of natural language inference examples that represent a range of semantic phenomena. Another example is the General Language Understanding Evaluation (GLUE; \citeNP{W18-5446}), a benchmark for a number of different natural language understanding tasks with training and test data from different domains and an evaluation platform. GLUE also provides a diagnostic dataset including manually-annotated examples that cover a range of linguistic phenomena. 

\citeA{W18-5454} present a study that uses such a resource, a dataset based on three reasoning tasks that are inspired by theory-of-mind experiments on children \citep{nematzadeh2018evaluating}. They use this dataset to test a memory network model on a question answering task and evaluate its reasoning capacity and generalizability.
Similarly, \citeA{W18-5430} use an existing dataset of semantic properties and augment it with negative examples for each property, which they use to examine which semantic properties are incorporated in word embeddings. They train supervised neural classifiers to identify each semantic property in a word embedding, and compare their performance to another model that uses vector cosine similarity. Their preliminary results suggest that interaction-based properties are captured better by neural models, whereas perceptual properties are not. 
\vskip 0.3cm

Overall, these studies vary in their objective and the type and scale of the resource they create and/or use, but there is no doubt that identifying various linguistic phenomena and challenges and providing carefully annotated test cases for each is the first step towards standardizing the evaluation of the new generation of analysis techniques, a point we will come back to in the last section.

\section{Analysis techniques} %

Many of the papers submitted to the workshop proposed new methods for
analysis of neural representations or applied existing methods to
particular models and data. Below we review the main families of
analytical methods represented at the workshop.

\subsection{Diagnostic auxiliary models} %
\label{sec:diagnostic}
A commonly applied technique consists in using neural representations as features, and training a
predictive model on top to predict information of interest such as
particular linguistic features. If the model is able to predict this
information with high accuracy, the inference is that the neural
representation encodes it. It is variously known as 
\begin{itemize}
\item auxiliary task \citep{adi2016fine},
\item diagnostic classifier \citep{hupkes2018visualisation}, 
\item probing \citep{P18-1198} or, 
\item decoding \citep{alishahi2017encoding}
\end{itemize}
The submission of \citeA{W18-5426} was
especially illustrative of the focus of the BlackboxNLP workshop, by
looking {\it under the hood} of neural language models, and it
received the best paper award. The authors apply the {\it diagnostic
classifier} approach to answer the question of how these models keep track
of subject-verb agreement in English. They decode number information
from internal states of a language model and analyze how, when, and
where this information is represented. Additionally they use the
mistakes of the diagnostic classifier to nudge the hidden states of
the network in the direction which would make the classifier more
likely to be correct and show that this intervention also improves the
results on the language modeling task.

\citeA{W18-5448,zhang2018language} use the diagnostic classifier
framework to investigate which primary training tasks are best suited
for transfer learning of syntactic information. They decode POS and
CCG tags from hidden layers of recurrent models trained on language
modeling, translation, auto-encoding, and on the skip-thought objective
\citep{kiros2015skip}. They find that the language modeling task learns
representations which overall encode syntax best; another interesting finding
is that randomly initialized LSTM states enable quite a high accuracy
of diagnostic POS and CCG tag classification: they ascribe it to the fact that
the hidden states of these random networks encode the identity of
words around the current position. This result suggests that successful
diagnostic classification should be interpreted with caution and not
automatically taken as proof that a particular neural model has
{\it learned} to encode syntax: it could be that the diagnostic classifier
itself is doing the learning.

\citeA{W18-5405} propose a method for evaluating
text encoders which is an interesting variation on the basic idea of
diagnostic auxiliary model. It is different from the canonical diagnostic
classifier setup in that the diagnostic model in this case is not a simple
classifier but rather a GAN-based image generation model which
attempts to produce an image given a representation of its caption. The
intrinsic quality of those images is evaluated using a variant of the Wasserstein distance with an independently trained critic. 
A similar metric with the critic modified to also receive the textual representations is then used to assess  the alignment of the images with the
text. 
The authors apply the
technique to a dataset of captioned medical X-rays and evaluate
several simple text encoding approaches.

\subsubsection{Correlation Analysis}

\citeA{W18-5438,saphra2018understanding} introduce an original twist on the idea of an auxiliary model. Instead of using a diagnostic classifier to probe learned neural language model representations, they train parallel recurrent models to do POS, semantic and topic tagging. They then measure the correlation between the activations of the language model and those of the tagger, using the technique of Singular Value Canonical Correlation Analysis (SVCCA) \citep{raghu2017svcca}. They use this method to examine how representations evolve over time and find that POS tag information is acquired before semantic or topics information.

\subsection{Nearest Neighbors and Kernels}
A natural approach to analyzing neural representations is based on
retrieving neighboring exemplars in the induced vector space. The
application of kernel methods is a related approach, in that kernels
are functions which encode some notion of pairwise similarity between
exemplars.

\citeA{W18-5416} combine two classes of techniques
commonly used in neural model interpretation: nearest neighbors and
sensitivity analysis. The authors apply the Deep k-Nearest Neighbor
model \citep{papernot2018deep} to perform neural text classification in
an interpretable fashion. The Deep k-Nearest Neighbor model classifies
a test input based on the labels of the training examples which are
nearest to it in representation space, as an alternative to the
standard softmax method.  Wallace et al.\ derive a metric of model
uncertainty, {\it conformity} based on this exemplar-based
classification, and compute input feature importance as a change in
conformity when a feature is removed (features in this case are
words). They show that conformity as a metric of uncertainty leads to
feature importance assignments which better agree with human
perception compared to using the confidence of a softmax-based
classifier. They also demonstrate that the method confirms some known
artifacts in the SNLI dataset.

\citeA{W18-5455,madhyastha2018end} investigate
the hypothesis that image captioning systems work by exploiting
similarity in multimodal feature space, i.e.\ essentially they
retrieve captions of training images close to a given test image in
representation space, and assemble a caption for the given image by
re-using elements of the captions of the retrieved images. The authors
support their conjecture by holding the caption generation portion of
the captioning system constant while varying image representations. It
turns out that performance is in many cases comparable across
substantially different image representations. This may be interpreted
as evidence against the fact that the specifics of these
representations play a major role in generating the captions, as long
as the representations encode a reasonable notion of image similarity.

\citeA{W18-5403} focus on explainability of neural
classifiers by providing the user with examples which motivate the
decision. They adapt the technique of {\it layerwise relevance
  propagation} \citep{bach2015pixel} to Kernel-based Deep Architectures
\citep{croce2017deep} in order to retrieve such examples. In essence,
in this architecture a vectorial input for a given structured symbolic
input (such as a parse tree) is built based on kernel evaluations
between the input and a subset of training examples known as
landmarks. With layerwise relevance propagation, the network decision
can be traced back to the landmarks which had most influence on
it. Based on this technique the authors build variants of explanatory
models for question classification and semantic role argument
classification. The models provide human-readable justification for their
decisions which were submitted for evaluation to human
annotators. Additionally, qualitative evaluation shows that the
explanations were able to capture semantic and syntactic relations
among inputs and landmarks.

\subsection{Saliency and attention}

Another well-represented approach to neural network analysis was 
the visualization of feature saliency and attention
maps. Feature saliency refers to the importance of input features in
influencing the output of a neural model. For models dealing with
written language, these would typically be words, n-grams or similar
patterns. Attention mechanisms provide a model-internal way to extract
similar information. Attention visualization may be especially
relevant for translation, where links between input and output
patterns can be highlighted. 

These type of approaches are often grouped
into two loose families:
\begin{itemize}
\item The patterns can be used to
illuminate a particular decision made by a model; this has been called
{\it prediction interpretability} \citep{W18-5408} or {\it local
introspection} \citep{W18-5421}. 
\item Typical or canonical
patterns which tend to lead to a particular decision can be extracted
and used to provide an explanation of what the model has learned in
general; this has been called {\it model interpretability} or {\it global
introspection}.
\end{itemize}
Approaches focusing on both prediction and model interpretability were represented, applied to convolutional as well as recurrent networks.

\subsubsection{Saliency  in Convolutional Networks}
The submission of \citeA{W18-5408} is an illustrative example of the
use of feature saliency mapping. The work deals with convolutional
neural models for text classification and refines common assumptions
about how CNNs work on discrete sequences. Specifically, they show that
max-pooling induces a thresholding behavior such that pattern activations
below a certain value do not affect prediction, that a single
filter often acts as a detector for several semantic classes of
n-grams, and that filters can also detect negative patterns. They show
application of their findings to  model interpretability and prediction
  interpretability.

\subsubsection{Saliency in Recurrent Networks}

\citeA{W18-5421} tackle the audio domain which has some inherent
difficulties for interpretability: while humans find it easy to
understand visual input patterns such as portions of images for vision
or n-grams for written language, the input used in speech recognition
such as waveforms or even spectrograms may be relatively hard to
interpret even for domain experts. In this study they analyze the
activation patterns of a fully convolutional model which transcribe
speech represented as spectrograms into character sequences. They
apply techniques previously found useful in interpreting image
classification. Firstly, for global introspection they apply
regularized activation maximization which finds a synthetic stimulus
which maximally activates a particular class (in this case a
letter). They also introduce a technique which averages and normalizes
all spectrogram frames predicted as the same letter.  Secondly, for
local introspection they experiment with layerwise relevance
propagation \citep{montavon2017explaining} and sensitivity analysis
\citep{gevrey2003review}. From qualitative analysis the authors
conclude that local introspection rarely results in interpretable
patterns. For global introspection, the aligned frame averaging
technique yields somewhat interpretable letter-specific patterns.

\citeA{W18-5418} discover salient patterns and use them for
interpreting the representations and decisions of a custom neural
architecture named Connectionist Bidirectional RNN
\citep{vu2016combining} applied to the classification of relations such as {\sc Cause-Effect, Content-Container} etc. \citep{hendrickx2009semeval}.
They
introduce two visualization techniques: firstly, they plot the
network's cumulative prediction score as a function of the position in
the sentence such that steep changes in the score correspond to
salient words. Secondly, they aggregate the patterns corresponding to
individual examples of a single given class to visualize the set of common
patterns which lead the network to predict this class. They carry out
a qualitative analysis of the discovered salient n-grams.

\citeA{W18-5443,Verwimp_2018} track how long information is retained
in the states of an LSTM language model by computing and decomposing
the gradient matrix of the state with respect to input word
embeddings, with a certain delay. The authors observe retention of input for
up to 30 steps, with increased selectivity for longer delays; they also
see some word classes, such as pronouns, retained for longer than others.

\citeA{W18-5428} also tackle the LSTM language model and discover
interpretable input patterns by applying HDBSCAN clustering
\citep{campello2013density} to the model's hidden
states. The technique reveals interpretable clusters which display
both character-level and grammatical structure patterns. Inspired by
the patterns discovered, the authors propose a method for extracting
word-embeddings from a character-level model, showcasing a practical
application of model analysis.

\citeA{W18-5437} aim to assign interpretable synthetic rather than
corpus-attested input patterns to individual dimensions (neurons) in
neural representations. The challenge with textual data is the
discrete nature of the input: they use the Gumbel softmax trick
\citep{jang2016categorical} to generate word sequences which maximize
activations for particular neurons. They apply this method to the
Imaginet architecture of \citeA{chrupala2015learning} and confirm one
of the findings in \citeA{kadar2017representation}: that the language model
part of the Imaginet architecture is more sensitive to function words than the visual
part, which tends to ignore them.  They also carry out a separate
quantitative evaluation of the synthetic patterns vs corpus-attested
in terms of achieved maximum activation.

\subsubsection{Attention mechanisms}

Two submissions \citep{W18-5431,W18-5444} analyze the internal workings of the encoder
component of the Transformer model \citep{vaswani2017attention}. The
Transformer is not autoregressive; instead it uses an encoding of
absolute word position as well as a number of (self-)attention heads
to encode structural information. This model has recently become a
state-of-the-art neural architecture for translation. Given its
novelty, not much is known about what exactly enables its superior
performance. \citeA{W18-5431} apply two main techniques:
deriving dependency relations from the attention weights, and diagnostic
classification of syntactic and semantic tags. The highlights of their
findings are: firstly, a significant amount of syntactic information
is encoded at each layer, and secondly, semantic information is easier
to decode from higher layers. Additionally they demonstrate the
transferability of the encoder weights from low to high resource
language pairs.
\citeA{W18-5444} describe methods to induce binary constituency trees as well as undirected dependency
trees from attention weights of the transformer. They observe that NPs and short VPs are often detected, unlike larger constituents, and that the induced dependency trees contain many flat treelets.

\section{Explainable architectures}

Another trend observed in the workshop was to use neural network architectures that are easier to analyze and understand. Such architectures are often more explicit in the types of linguistic representations they use; for example, they use latent variables for incorporating linguistic information such as lexical categories or tree structures, or are specifically trained to learn structured output. Alternatively, specialized attention mechanisms are employed which provide more insight into the linguistic features that models rely on for performing their final task.

\subsection{Latent variables for capturing linguistic representations}

In his invited talk, Graham Neubig presented a series of works on developing more explainable architectures that can be trained in an unsupervised or semi-supervised fashion, and allow for the emergence of interpretable linguistic structure.  Specifically, \citeA{zhou17acl} propose an architecture called {\it multi-space variational encoder-decoder} which uses discrete (e.g. POS, tense, person) and continuous (e.g. lemma) latent variables for transforming labeled sequences in a supervised or semi-supervised fashion. In a similar approach, \citeA{yin18acl} propose a variational auto-encoding model for semantic parsing which learns meaning representations in unlabeled data as tree-structured latent variables. In both these cases the inclusion of the latent variables improves the performance in the down-stream tasks, but equally importantly, such structured variables make it easier to analyze the linguistic representations that models extract from labeled and unlabeled data.

\citeA{W18-5450} also suggest sparse and structured latent computation as a mechanism for improving interpretability in neural networks. Their goal is to use various techniques for transforming the dense inner representations of neural networks into sparse representations, which show what part of input the model bases its decision on. Various techniques are suggested for this purpose, for example using regularization techniques in order to incorporate prior assumptions  \citep{niculae2017regularized}, or  learning latent structure predictors such as parses or aligners \citep{niculae2018sparsemap}. \citeA{W18-5422} apply similar techniques to sentence embeddings in order to make them more sparse and therefore easier to interpret. 

For tasks such as sentiment analysis where negations and contrastive phrases change the polarity of the sentence, it is more informative to interpret the task as a series of incremental inferences. \citeA{W18-5427} propose a special attention mechanism which makes such interpretation possible. Their iterative recursive attention model relies on   a non-linear transformation of the representations from previous steps to build a recursive representation of the input sequence in an incremental manner.

\subsection{Generating interpretable output}

As an attempt to improve the {\it global introspection} of a trained model and to provide explicit explanations for its output, \citeA{W18-5411} propose a technique for generating a set of if-then-else rules that explain the prediction of a class label based on the most important features extracted from the input. They do this by first analyzing the feature weights and highlighting the most informative ones in a trained model, then mapping them to a set of discrete features that represent either a positive, a negative, or no correlation with a class label. They induce a set of rules from this reduced feature space that best explains the model’s predictions. 

\citeA{W18-5420} generate interpretable output by training a neural machine translation model to incorporate explicit word alignment information in the representation of the target sentence: the model is trained to generate a target sentence in parallel with its alignment with the source sentence (as a linear sequence of operations).

\citeA{W18-5434} borrow explanation techniques from Computer Vision called PatternNet and PatternAttribution \citep{kindermans2017learning}, which estimate the signal and the noise in the input to a model. They apply these techniques to a CNN trained for text classification, and retrieve neuron-wise signal contributions in the input vector space. They align the aggregated contribution scores with the input text and show that the model especially pays attention to bigrams, and ignores stop words.

\section{Linguistics and formal language theory}

\subsection{What do neural networks learn about language?}
\label{sec:behavioral_linguistics}

A number of papers in the workshop sought to obtain an interpretable characterization of the linguistic knowledge captured by a neural network by studying the network's behavior on examples that illustrate particular linguistic phenomena. This approach often leverages experimental paradigms from psycholinguistics that were originally developed to characterize the representations used by humans.

\subsubsection{Agreement as a probe into structural representations}
\citeA{W18-5412} trained RNNs to predict the agreement features of a verb in Basque; perfect accuracy on this task requires identifying the subject of the verb, which in turn requires sophisticated syntactic representations \citep{linzen2016assessing}. In Basque, which differs from English in a large number of properties, accuracy was substantially lower than in earlier studies on English. While the difference may be due to a number of factors, an intriguing possibility is that the inductive biases of popular neural architectures are better suited to English than to other languages. 

\citeA{W18-5453} test whether syntactic representations in a language model can transfer across languages. They trained an RNN LM on a corpus formed by concatenating a large French corpus and a small Italian corpus, and evaluated the model on the Italian agreement dependencies collected by \citeA{gulordava2018colorless}. They found that adding the large French corpus provides a modest improvement over training on the smaller Italian corpus alone, but the bilingual model is less effective at resolving agreement dependencies than a model trained on a monolingual Italian corpus of a matched size, suggesting that cross-linguistic transfer is limited even across related languages.

\subsubsection{Structural dependencies beyond agreement}
A number of studies investigated whether RNN LMs show awareness of linguistic phenomena beyond subject-verb agreement; studying these more complex phenomena often requires more experimental ingenuity than subject-verb agreement does. \citeA{W18-5424} investigate whether an RNN LM shows sensitivity to the restrictions on the distribution of negative polarity items (NPIs), such as \textit{any}. NPIs can only occur in a limited set of licensing environments \citep{giannakidou2011negative}; for example, \textit{I did not eat any cookies} is grammatical, but \textit{*I ate any cookies} (perhaps with an intended meaning similar to \textit{I ate some cookies}) is not. They found that the model was able to detect the environments in which NPIs are allowed to appear; however, this ability degraded with the distance between the NPI and the word that signals the licensing environment.

\citeA{wilcox} study the representation of filler-gap dependencies in RNN LMs. As an illustration of this phenomenon, consider the sentence \textit{I know what the lion devoured $\rule{0.7cm}{0.15mm}$ at sunrise}. The verb \textit{devoured} typically requires a direct object, but this sentence does not have an object; this object can be thought of as having been displaced to the left of \textit{the lion} and replaced with the question word \textit{what}. \citeA{wilcox} found that two RNN LMs showed significant awareness of such gaps, as well as of some (but not all) of the restrictions on their distribution (``island constraints''; \citeNP{ross1967constraints}). 

\citeA{W18-5409} analyzed bidirectional RNNs trained to resolve the antecedent of a ``sluiced'' question word, as in \textit{If this is not practical, explain why $\rule{0.7cm}{0.15mm}$}; here the phrase \textit{this is not practical} needs to be interpreted again following the question word, even though it does not reappear. They report both behavioral analyses on challenging cases and analyses of the activations of the network; for example, they show that the distance between the embedding of the question word (\textit{why}) and the network's activation is minimal at the edge of the antecedent, and conclude that the network reactivates the question word when it encounters the antecedent (recall that the network processes the sentence both from right to left and from left to right).

\subsubsection{Other phenomena}
\citeA{W18-5440} propose a new diagnostic task for measuring the syntactic sophistication of vector sentence representations generated by neural networks. The task measures whether the tense of the main clause verb can be decoded using a linear classifier (following \citeNP{P18-1198}; see also section~\ref{sec:diagnostic}). \citeauthor{W18-5440} observe that correctly identifying the main clause verb can require sophisticated syntactic representations when the sentence includes multiple clauses. In particular, they sample sentences in four languages that have multiple verbs that conflict in their tense; for example, the main verb in \textit{we know who won} is in the present tense (\textit{know}), but the past-tense embedded clause verb \textit{won} serves as a distractor. This approach, in which test sentences are selected based on their linguistic complexity instead of being sampled at random from the corpus, is typical of the linguistically-informed approach to neural network analysis \citep{linzen2019what}.

\citeA{W18-5432} propose to evaluate the grammaticality of the output of a neural machine translation model using a precision HPSG grammar; unlikely standard parsers, which may produce a parse for ungrammatical English sentence, a parser based on a precision grammar can be used to determine if a sentence is grammatical in English. They find that the majority of the target sentences in a French-to-English translation dataset were parseable.  A preliminary error analysis of the unparseable sentences shows that about 20\% of them contained agreement errors; future analysis of other types of errors can provide insight into the limitations of the language model learned by the decoder.

By contrast with the preceding studies, which were concerned with the syntactic awareness of models in which words were the basic input unit, \citeA{W18-5417} investigate the morphosyntactic generalizations acquired by character-based LMs, which often do not have explicit representations of words. Qualitatively, they show that samples from a character-based model trained on English text include real words, and pseudowords that made up of real English morphemes (such as \textit{indicatements} or \textit{breaked}), but also nonwords that do not conform to English phonotactics (\textit{ouctromor}). Perhaps surprisingly, they discovered a highly interpretable recurrent unit, which recognizes the ends of words and sub-word units. They use diagnostic tasks to determine whether the model's internal representations contain sufficient information for morphological segmentation and part-of-speech tagging. Finally, they argue that the model shows some sensitivity to morpheme's selectional restrictions (e.g., \textit{-ment} is more likely to attach to a verb than to other parts of speech). At the same time, there appears to be a limit to the amount of information that can be gleaned by a character RNN; \citeA{W18-5447} show that explicit morphological case annotation improve the performance of a dependency parser compared to a model trained only on characters.

\subsection{Synthetic languages}
\label{sec:synthetic}

Synthetic training data, which can be carefully controlled, can illuminate the learning capabilities of an architecture (or, more precisely, the combination of an architecture and a learning algorithm). Networks trained on synthetic data can be easier to analyze, and more extensive experiments can be run without large computational power---indeed, in the early days of neural networks, those were the only experiments that were feasible \citep{elman1990finding,hochreiter1997long}. If an architecture is unable to learn a particular phenomenon given a large amount of synthetic data, we may be justified in being pessimistic about its ability to learn it with natural language. Of course, the converse is not always true: the inductive bias of the architecture may be sufficient for learning from the distribution of constructions in a particular synthetic data set, but not from the distribution of natural language, where critical cases that disambiguate multiple hypotheses about the language may be rare. 

In his invited talk, Yoav Goldberg discussed a study in this vein. It is a well-known theoretical result that RNNs are Turing complete \citep{siegelmann1995computational}. This result relies on the assumption that the system can process infinite precision numbers; in practice, of course, all neural networks are implemented in finite-precision computers. \citeA{weiss2018practical} show that this implementational fact has important consequences.  In particular, fine-grained differences between GRUs and LSTMs in the details of the recurrence equations  lead to large differences in their ability to count;\footnote{Specifically, a GRU cell's memory is constrained to be between $-1$ and $1$, whereas an LSTM is not constrained in that way} such an ability to count is essential for recognizing formal languages such as $a^nb^n$, and may be relevant for the processing of recursive embedding in natural language. This result, obtained using a synthetic language, is particularly striking because GRUs and LSTMs are not typically considered to achieve substantially different performance in applications.

Two groups trained RNNs with different recurrent units to recognize Dyck languages, which are simple context-free language that requires matching opening and closing brackets \citep{W18-5414,W18-5425}. The studies find that the RNNs are able to deal with shorter strings, but generalize poorly to strings that are longer than those on which they were trained. \citeA{W18-5414} conclude from their results that LSTMs' success in practical applications relies on approximations of natural language grammar, which may work well in most practical cases, but do not conform to the theoretical analysis of the grammar of the language (which is at least context-free for most natural languages).

Going beyond the classic test of recognizing whether a string is a part of a particular formal language, a number of papers instead trained networks to assign an interpretation to a string in the language. \citeA{W18-5456} reports that LSTMs are able to learn compositional interpretation rules, but only if their order of application follows the sequential order of the sentence, and they require extensive training to learn to do so. Likewise, sequence-to-sequence LSTMs trained to translate commands to sequences of actions on the SCAN dataset \citep{lake2018generalization} do not learn systematic compositional rules \citep{W18-5413}. Finally, \citeA{W18-5407} propose a modification to the SCAN dataset that makes the task even more challenging and exposes poor generalization performance in sequence-to-sequence networks. Taken together, these synthetic language studies suggest that RNNs, while often very effective for cases that are similar to those they were trained on, do not generalize in the way that a human would.

\subsection{Injecting linguistic knowledge into neural networks}

Classic language technologies are based on symbolic representations that are readily interpretable to a human with basic training in linguistics; such symbolic representations include constituency parses, logical formulas or meaning graphs. A number of studies explored ways to reincorporate computational elements from linguistics or formal language theory into neural network architectures; in addition to their potential to improve interpretability, such hybrid neural networks may have inductive biases that are more appropriate to language learning than are the biases of sequential RNNs (see section \ref{sec:synthetic}). 

\citeA{W18-5433} analyze the behavior of one such architecture: stack-augmented neural networks \citep{grefenstette2015learning}, an architecture motivated by the observation that adding a stack to a finite automaton enables it to recognize context-free languages \citep{chomsky1962context}. An attractive feature of this architecture is that its utilization of the stack can be visualized in a straightforward and interpretable way. Along with interpretable visualizations for a number of synthetic languages, \citeauthor{W18-5433} also find that the external stack is underutilized when the controller is an LSTM rather than a simpler neural network. This case study shows that it is not sufficient to add a linguistically motivated component to the architecture; it is crucial to ensure that the model cannot circumvent that component.

A number of recently proposed architectures incorporate constituency parses into neural networks: words are composed based on the parse, rather than from left to right as in a standard RNN. While early architectures required the parse to be provided as part of the input, more recent architectures learn to parse sentences only using supervision from a downstream task such as sentiment analysis, without explicit supervision in the form of parse trees. Prior to the workshop, \citeA{williams2018latent} showed that the trees induced by some of these models did not conform to linguistic intuitions and were inconsistent across run. By contrast, in an abstract presented at the workshop, \citeA{W18-5452} analyzed the trees produced by one of these grammar-induction networks---the Parsing-Reading-Predict Network \citep{shen2018neural}---and found that these trees were much more consistent with gold parses than the trees produced by previous models. A notable feature of this work is that it is a replication study that addresses issues with the experimental design of the paper that originally proposed the model.

\section{Future trends and outlook}
\label{sec:future}
This first edition of the BlackboxNLP workshop brought together a
large amount of recent work on issues related to the analysis and
interpretability of neural models, and thus allows for a preliminary
assessment of the trends and potential points of focus for the future.

\paragraph{Evaluation}
One point which stood out to us is that there is no consensus on the
best way of evaluating the different analytical techniques which are
being introduced. A number of submissions resorted to
qualitative evaluation to see whether the conclusions reached via a
particular approach have face validity and match pre-existing
intuitions. While this is often a necessary and helpful first step we
believe that going forward it needs to be supplemented by more
rigorous quantitative evaluation in order for the field as a whole to
make measurable progress and become accepted as part of mainstream
NLP. Some papers carried out quantitative evaluation of explanations
of model decisions by human annotators. As currently practiced it can
have its own issues. Specifically, when an explanation matches what a
human would see as a reasonable basis of a particular decision, it
does not necessarily follow that this was the basis that caused the
model to make this decision. We see developing agreed-upon approaches
to the evaluation of analytical and explanatory techniques as a
major challenge for the field in the immediate future.

\paragraph{Benchmarks}
A related issue concerns datasets and benchmarks. Some popular
datasets for evaluation have already emerged such as the agreement
datasets \citep{linzen2016assessing,gulordava2018colorless}. As we have
seen in Sections~\ref{sec:dataset} and~\ref{sec:behavioral_linguistics} the workshop has seen some
more contributions in that space. In future we would also like to see
formal shared or unshared tasks to facilitate further progress. Again,
we see further developments and gradual standardization here as
crucial for the field.

\paragraph{Neuroscience}
We saw several disciplinary traditions represented at the workshop,
including NLP, computer vision, speech processing, formal linguistics
and psycholinguistics. The workshop also featured an inspiring invited
talk on modeling language representations in the human brain with
neural network models by Leila Wehbe. In future we would be very
interested in welcoming neuroscientists to join the
BlackboxNLP community in larger numbers and collaborate on answering fundamental
questions about language in human and artificial brains.

\bibliographystyle{chicago}
\bibliography{blackbox_jnle,analysis}
\end{document}